\RequirePackage{fix-cm}
\documentclass[11pt]{article}

\usepackage[final]{acl}

\usepackage{times}
\usepackage{latexsym}
\usepackage[T1]{fontenc}
\usepackage[utf8]{inputenc}
\usepackage{microtype}
\usepackage{inconsolata}
\usepackage{graphicx}
\usepackage{booktabs}
\usepackage{adjustbox}
\usepackage{amsmath}
\usepackage{amssymb}

\title{DisasterRoute: Evidence-Guided Dual-System Routing for Visual Disaster Question Answering}

\author{
  Tan Zhang$^{1}$ \quad
  Xing Deng$^{1}$ \quad
  Lu Zhang$^{2}$  \quad
  Xiaofeng Zhu$^{3}$ \quad
  Ping Hu$^{1}$ \\
  $^{1}$UESTC~~~~~~$^{2}$DLUT~~~~~~$^{3}$Hainan University
}

\begin{document}
\maketitle

\begin{abstract}
Visual question answering from low-altitude UAV imagery must deliver timely and accurate judgments to support disaster response under limited edge computing resources. Routine questions may be answered through direct decisions, whereas more challenging questions may benefit from step-by-step reasoning. However, existing research primarily focuses on improving models' visual understanding and generative reasoning capabilities, with limited exploration of efficient direct decision-making and its adaptive coordination with generative reasoning. We develop Disaster-S1 based on a vision-language model adapted to the disaster domain. Trained for option discrimination, it outputs a probability distribution over answer options in a single multimodal forward pass for fast decision-making. We further propose DisasterRoute, a dual-system routing framework that combines prediction confidence with disaster evidence to decide whether to accept the initial answer or invoke the disaster reasoning model DisasterVL for review. By allocating inference computation on demand, the framework retains the efficiency of direct decision-making and improves overall question-answering accuracy through selective review. Experiments on DisasterBench show that Disaster-S1 achieves superior accuracy with low latency, while DisasterRoute further improves accuracy through selective review at substantially lower latency than full reasoning.
\end{abstract}

\begin{figure}[t]
    \centering
    \includegraphics[width=\columnwidth]{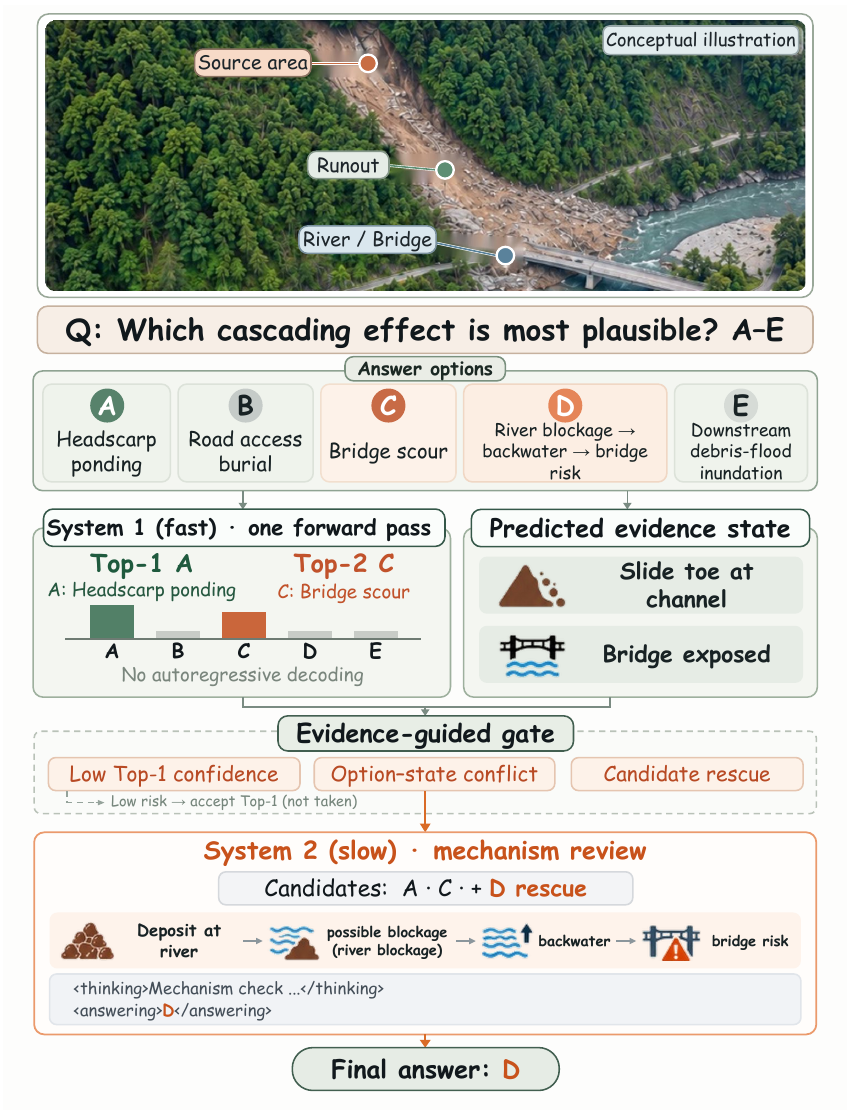}
    \caption{Geological source--motion--impact cues guide selective inference in DisasterRoute. Disaster-S1 makes one-pass option decisions, while DisasterVL reviews questions with uncertain or conflicting evidence.}
    \label{fig:teaser}
    % \vspace{-0.5cm}
\end{figure}

\section{Introduction}

Effective disaster response depends on timely and accurate situational assessments. Responders use on-site observations to identify disaster conditions, anticipate developments, and assess threats to people, infrastructure, and access routes. These judgments must promptly inform response actions, making latency important alongside accuracy. Low-altitude unmanned aerial vehicles (UAVs) can reach hazardous or inaccessible locations and provide timely, close-range visual information. However, occlusion, viewpoint changes, and limited coverage can prevent a single image from fully capturing the hazard source, propagation path, and impact zone, leaving different judgments with different evidence~\cite{Zhang2026DisasterBench,Dille2026HazardContinuum}. Meanwhile, on-site applications face limited edge computing resources and tight response deadlines, while sustained costly inference adds computational and time demands~\cite{Kyrkou2020EmergencyNet}. Visual disaster question answering must therefore balance accuracy with computational and time costs.

Multimodal large language models have extended remote-sensing and disaster analysis beyond scene recognition and description toward interactive understanding and reasoning~\cite{Kuckreja2024GeoChat,Soni2025EarthDial,Wang2025DisasterM3}. DisasterBench provides visual question answering tasks spanning multiple disaster stages in low-altitude UAV imagery, while the domain-adapted DisasterVL demonstrates lightweight models' potential for disaster understanding and generative reasoning~\cite{Zhang2026DisasterBench}. These advances provide domain knowledge and reasoning capabilities, but inference procedures still affect response costs. Multiple-choice questions ultimately require selecting among candidate answers. Autoregressive answering generates outputs sequentially and incurs decoding costs even with lightweight models~\cite{Leviathan2023SpeculativeDecoding}. This motivates investigating option-discrimination training to strengthen existing disaster-domain models' direct decisions and enable efficient answering through candidate scoring. In visual disaster question answering based on low-altitude UAV imagery, research has focused mainly on visual understanding, generative reasoning, and domain adaptation. Efficient direct decision-making and its on-demand coordination with generative analysis remain underexplored.

We argue that visual disaster question answering should coordinate fast decisions and further analysis according to question requirements and available evidence. Routine questions may not require complete step-by-step deliberation, whereas harder questions may require comparing competing explanations that an initial judgment alone cannot resolve. Drawing on the distinction between fast and slow reasoning~\cite{Evans2013DualProcess,Saha2025System1x,Sun2025FastSlow}, System 1 first forms an initial judgment through direct option scoring. A subsequent decision determines whether System 2 performs generative review, concentrating further reasoning on selected cases. For example, a visible deposition zone may support impact assessment while remaining insufficient to identify the initial trigger. Confidence-related signals help assess answer reliability in selective visual question answering~\cite{Whitehead2022ReliableVQA,Dancette2023SelectiveVQA}. However, a model's preference among options does not by itself establish whether the image contains the evidence required by the question. Review selection should also consider question-specific evidence sufficiency and agreement between candidate explanations and scene state. Restricted candidate sets must also preserve opportunities to correct initial predictions by avoiding premature exclusion of the correct answer.

To this end, we build Disaster-S1 on a disaster-domain-adapted vision--language model~\cite{Zhang2026DisasterBench} and strengthen direct option decisions through option-discrimination training. During inference, Disaster-S1 produces a probability distribution over five options in a single multimodal forward pass, without autoregressive generation. We further propose DisasterRoute, a dual-system routing framework. After obtaining Disaster-S1's initial decision, the framework combines prediction confidence, estimated question-specific evidence sufficiency, and option agreement with the predicted scene state to decide whether to accept the answer directly or invoke the disaster reasoning model DisasterVL for review. As the conceptual example in Figure~\ref{fig:teaser} illustrates, DisasterVL compares the two highest-ranked options using the original image. When rescue conditions are met, one candidate outside the Top-2 is added to reduce the risk of omitting the correct answer. Disaster-S1 provides efficient direct decisions, while DisasterRoute improves overall accuracy through selective review. Together, they allocate computation on demand, retaining a fast path for most questions and concentrating additional analysis on selected cases to balance accuracy and response efficiency.

Our contributions are as follows:
\begin{itemize}
    \item We develop Disaster-S1 for visual disaster question answering. Option-discrimination training strengthens direct decision-making, enabling option probability distributions from a single multimodal forward pass without autoregressive generation.
    \item We propose DisasterRoute, a dual-system routing framework combining prediction confidence and disaster evidence to coordinate direct decisions and on-demand generative review. Conditional candidate rescue reduces the risk of omitting the correct answer during review.
    \item We evaluate fast decisions and selective review on DisasterBench, examining accuracy, calibration, routing behavior, candidate coverage, and review outcomes. Latency, throughput, and GPU memory evaluations assess the benefits and resource costs of this coordination.
\end{itemize}

\section{Related Work}

\subsection{Multimodal Disaster Understanding}

Multimodal learning has extended remote-sensing analysis from specialized perception to visual--language interaction. GeoChat supports grounded dialogue and visual question answering for satellite imagery~\cite{Kuckreja2024GeoChat}; LHRS-Bot uses volunteered geographic information to improve domain alignment~\cite{Muhtar2025LHRSBot}; and EarthDial integrates multisensory and multitemporal observations in a conversational model~\cite{Soni2025EarthDial}. For disaster analysis, MONITRS links temporal observations with language for event monitoring~\cite{Revankar2025MONITRS}, RSCC studies bi-temporal change captioning~\cite{Chen2025RSCC}, and DisasterM3 provides multi-hazard, multi-sensor data for damage assessment and response~\cite{Wang2025DisasterM3}. DisasterBench further introduces low-altitude UAV questions spanning causal attribution, hazard propagation, damage analysis, and response decisions, together with the domain-adapted DisasterVL model~\cite{Zhang2026DisasterBench}. These works advance disaster understanding, but questions about visible impacts, initiating causes, and possible consequences can require different evidence and reasoning effort. We therefore study how to allocate computation across questions, using direct option decisions when the available evidence supports them and generative review when further examination is warranted.

\subsection{Jev-Style System 1 Decisions and Selective Review}

Dual-process theories distinguish fast judgments from more deliberate examination of alternatives~\cite{Evans2013DualProcess}. Jev adopts a System 1 decision interface that returns structured probabilities over predefined outputs without free-form generation~\cite{Almeida2026Jev}, and an independent evaluation has examined its performance and decision probabilities across diverse tasks~\cite{Deusser2026JevBenchmark}. Jev-style research has also extended direct decisions to visual inputs: Visual Jev shares image context across questions and reads candidate probabilities from a vision--language model's language-model head~\cite{Yu2026VisualJev}, while PixelJev accepts an image and a runtime candidate set to produce candidate-conditioned visual decisions~\cite{Zhou2026PixelJev}. LLM-as-Jev instead extracts categorical decisions from next-token probabilities and studies when fine-tuning is useful~\cite{Li2026LLMasJev}. Beyond decision accuracy, JEVal finds that general decision models can be less reliable when the available evidence is insufficient or uncertainty is poorly estimated~\cite{Duan2026JEVal}. Fast--slow planning and visual-agent studies show how direct decisions can be coordinated with more deliberative computation~\cite{Saha2025System1x,Sun2025FastSlow}. More directly, JEV-as-a-Judge escalates uncertain Jev verdicts to a reasoning judge~\cite{Li2026JevJudge}, while REFLEX uses Jev as a fast agent decision layer and calls a generative model when confidence is low or generation is required~\cite{Wu2026ReflexJev}. Reliable VQA likewise indicates that softmax confidence alone may be insufficient for deciding when to trust an answer~\cite{Whitehead2022ReliableVQA}. In UAV-based disaster question answering, Disaster-S1 makes a single-pass decision over five options, and DisasterRoute combines confidence with question-specific evidence sufficiency and option--state agreement to decide whether DisasterVL should review it. Conditional candidate rescue further addresses the possibility that the correct answer falls outside the initial Top-2 review set.

\begin{figure*}[t]
  \centering
  \includegraphics[width=\textwidth]{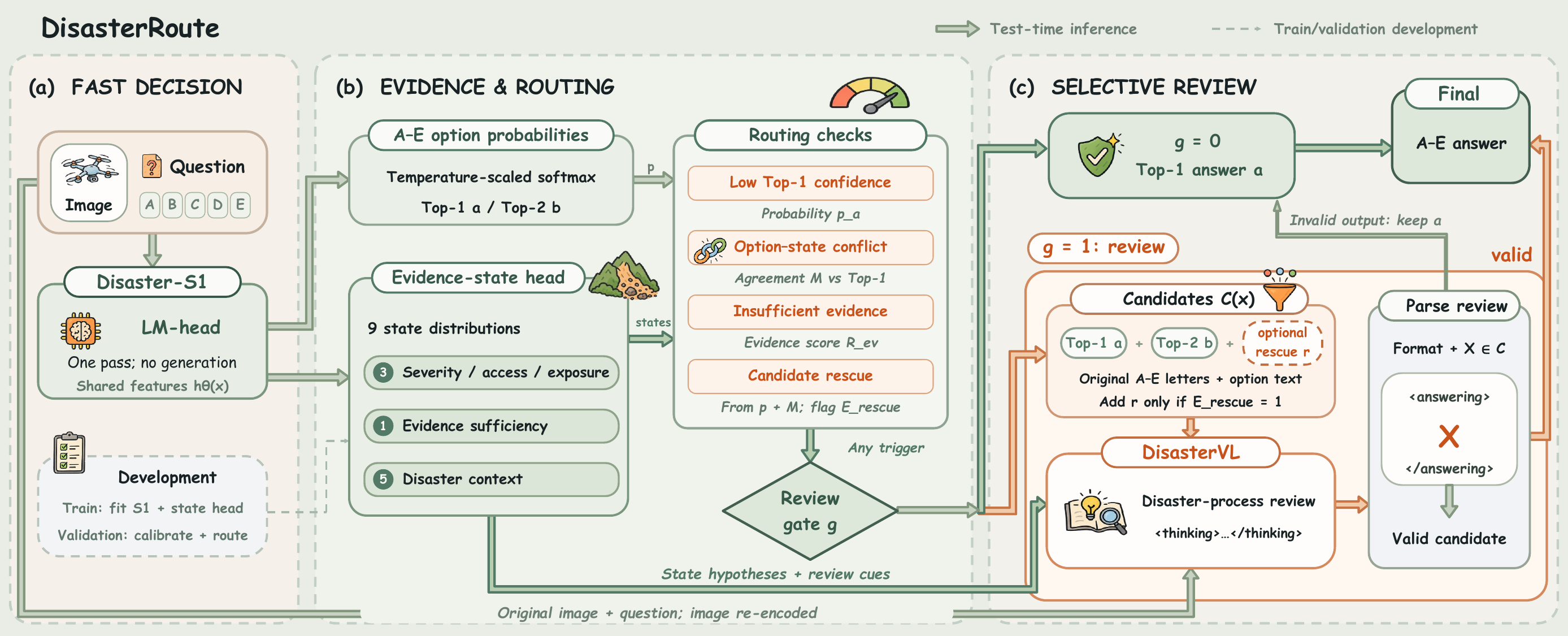}
  \caption{Overview of DisasterRoute. (a) Disaster-S1 scores A–E in one pass. (b) Shared features support evidence-state prediction and routing using confidence, option–state agreement, evidence sufficiency, and candidate rescue. (c) Direct cases retain Top-1, while routed cases invoke DisasterVL with the original image, question, and two or three original-letter candidates. Invalid or out-of-set outputs revert to Top-1. Dashed arrows denote development-only flows.}
  \label{fig:mach_overview}
\end{figure*}

\section{Methodology}
\label{sec:methodology}

Low-altitude UAV disaster question answering requires timely decisions from visual evidence that may be incomplete or unevenly distributed across a scene. Answer options can differ in their descriptions of disaster progression, severity, access conditions, or exposed objects. Consequently, a confident option prediction need not be supported by sufficient evidence, while generating a full reasoning trace for every question incurs unnecessary latency. To address this tension, we introduce DisasterRoute, which connects fast option selection, evidence-state assessment, and selective generative review in a unified pipeline. As shown in Figure~\ref{fig:mach_overview}, Disaster-S1 first produces a five-option decision in one forward pass; an evidence-state head and routing module then determine whether that decision should be accepted or reviewed; only selected questions are sent to DisasterVL for constrained verification.

Given an image $I$, a question $q$, and five options
$\mathcal O=\{o_k\mid k\in\mathcal K\}$ indexed by
$\mathcal K=\{A,B,C,D,E\}$, let $x=(I,q,\mathcal O)$ denote an input
instance. The system returns an original A--E option letter $\hat y$.
Disaster-S1 supplies the initial option ranking, while the evidence-state
head supplies structured, fallible hypotheses about the current
instance. A gate $g\in\{0,1\}$ either accepts the leading option or
invokes DisasterVL with a restricted candidate set. Model fitting and
cross-fitting use training data; calibration and routing parameters are
selected on validation data. All components are fixed before final
testing.

\subsection{System 1: Single-Pass Disaster Decision}
\label{sec:method_system1}

Figure~\ref{fig:mach_overview}(a) depicts the fast decision stage.
Disaster-S1 receives the image, question, and all five options. Rather
than autoregressively generating an answer, it reads the logits of the
five option letters at the answer position. Let $h_\theta(x)$ be the
image--text hidden representation at that position, $W$ the language
model output matrix, and $\nu(k)$ the token index of letter $k$. The
option logit is
\begin{equation}
    z_k(x)=[Wh_\theta(x)]_{\nu(k)},
    \qquad k\in\mathcal K.
    \label{eq:dr_option_logits}
\end{equation}

During training, these five logits are normalized to obtain
$\pi_k(x)=\exp(z_k(x))/
\sum_{j\in\mathcal K}\exp(z_j(x))$. For labeled training examples
$\{(x_i,y_i)\}_{i=1}^{N}$, Disaster-S1 minimizes the five-way
cross-entropy
\begin{equation}
    \mathcal L_{\mathrm{S1}}
    =-\frac{1}{N}\sum_{i=1}^{N}\log\pi_{y_i}(x_i).
    \label{eq:dr_system1_loss}
\end{equation}
At inference, a temperature $T>0$ selected on validation data yields
the option probabilities used for routing:
\begin{equation}
    p_k(x)=
    \frac{\exp(z_k(x)/T)}
         {\sum_{j\in\mathcal K}\exp(z_j(x)/T)}.
    \label{eq:dr_option_probabilities}
\end{equation}
Let $a$ and $b$ denote the highest- and second-highest-ranked options.
Temperature scaling preserves their ranking. Option $a$ is the initial
answer, $(a,b)$ forms the basic review candidate set, and $p_a$
provides a confidence signal. The same forward pass also supplies
$h_\theta(x)$ to the evidence-state head; state prediction does not
require another invocation of the vision--language backbone.

\subsection{Cross-Fitted Evidence States and Routing}
\label{sec:method_evidence_routing}

Figure~\ref{fig:mach_overview}(b) combines the initial option
distribution with nine evidence-state fields: hazard type, motion
pattern, spatial evidence, trigger visibility, secondary hazard,
exposed object, severity stage, access state, and evidence sufficiency.
Severity, access, and exposure can be compared with option text;
evidence sufficiency indicates whether the available observations
support the judgment requested by the question. The remaining fields
provide disaster-context cues for subsequent review. State outputs are
hypotheses that may be incorrect, rather than verified observations.

The evidence-state head $E_\eta$ reads the answer-position
representation $h_\theta(x)$. A shared transformation produces
$u(x)$, and a field-specific classifier outputs a distribution over
the legal classes $\mathcal V_f$ of each field $f$:
\begin{equation}
    \begin{aligned}
        u(x) &= f_\eta\!\left(h_\theta(x)\right),\\
        P_{\eta,f}(v\mid x)
        &= \operatorname{softmax}
           \!\left(W_f u(x)+b_f\right)_v,\\
        &\qquad v\in\mathcal V_f.
    \end{aligned}
    \label{eq:dr_state_head}
\end{equation}
Thus, the head returns nine separate categorical distributions, not
one probability per field. For each field, its predicted label and
associated confidence are
\begin{equation}
    \begin{aligned}
        s_f &= \arg\max_{v\in\mathcal V_f}
               P_{\eta,f}(v\mid x),\\
        c_f &= \max_{v\in\mathcal V_f}
               P_{\eta,f}(v\mid x).
    \end{aligned}
    \label{eq:dr_state_prediction}
\end{equation}

Supervision targets $\tilde{\mathbf s}_i$ are constructed only for
training examples from training-split disaster annotations,
image--question evidence annotations, and predefined option-semantic
rules. A training answer may be used to construct a supervision
target, but is never supplied as an input to the state head at
inference. In particular, evidence-sufficiency targets characterize
whether the image supports the requested judgment; they are not
defined merely by whether another model answered the question
correctly. The state head minimizes
\begin{equation}
    \begin{aligned}
        &\mathcal L_{\mathrm{state}}(\eta)\\
        &\quad=\frac{1}{N}\sum_{i=1}^{N}
         \sum_{f\in\mathcal F_{\mathrm{state}}}
         \lambda_f\,
         \mathrm{CE}\!\left(
             P_{\eta,f}(\cdot\mid x_i),
             \tilde s_{i,f}
         \right),
    \end{aligned}
    \label{eq:dr_state_loss}
\end{equation}
where $\mathcal F_{\mathrm{state}}$ contains the nine fields and
$\lambda_f$ controls the contribution of each field.

To obtain out-of-fold state predictions on the training split, we
partition its source images into $K$ folds, keeping all questions
associated with the same image in one fold. For fold $t$, a state head
$E_{\eta^{(-t)}}$ is trained using the other $K-1$ folds and predicts
the held-out fold:
\begin{equation}
    \widehat{\mathbf s}^{\,\mathrm{OOF}}_i
    =E_{\eta^{(-t)}}\!\left(h_\theta(x_i)\right),
    \qquad x_i\in\mathcal D_t.
    \label{eq:dr_crossfit}
\end{equation}
Each training example therefore receives a state prediction from a
head that was not trained on that example's state labels. These
out-of-fold predictions support development and assessment of
state-dependent routing signals without relying on the head's
in-sample predictions. After the configuration is fixed, a final head
$E_{\eta^\star}$ is fitted on the complete training split and used for
both validation and test inference. Validation data determine the
calibration and routing parameters; test labels do not participate in
state prediction or route construction.

Routing first compares the predicted state with each answer option.
Let
$\mathcal F_{\mathrm{cmp}}
=\{\mathrm{sev},\mathrm{acc},\mathrm{exp}\}$
denote severity, access, and exposure, and let $\psi_f(o_k)$ extract
the corresponding attribute from option $o_k$. The match value
$m_f(k)$ is $1$ for agreement, $0$ for disagreement, and $1/2$ if the
option does not specify that attribute. With field weight $w_f$ and
confidence floor $\epsilon>0$, option--state agreement is
\begin{equation}
    \begin{aligned}
        \omega_f &= w_f\max(c_f,\epsilon),\\
        M_k &=
        \frac{\sum_{f\in\mathcal F_{\mathrm{cmp}}}
                    \omega_f m_f(k)}
             {\sum_{f\in\mathcal F_{\mathrm{cmp}}}\omega_f}.
    \end{aligned}
    \label{eq:dr_option_state_agreement}
\end{equation}
The neutral match value prevents an unspecified attribute from being
treated as an explicit contradiction. $M_k$ measures compatibility
with the selected state fields, not the correctness of a complete
disaster-process explanation.

To combine initial option ranking with state agreement, we compute
\begin{equation}
    U_k=(1-\alpha)M_k+\alpha p_k,
    \qquad 0\leq\alpha\leq1.
    \label{eq:dr_candidate_support}
\end{equation}
Let $k_M=\arg\max_{k\in\mathcal K}M_k$. If $k_M\ne a$ and
$M_{k_M}\geq M_a+\delta_v$, another option agrees substantially more
strongly with the predicted state than the initial answer. The
evidence-sufficiency distribution additionally gives
\begin{equation}
    \begin{aligned}
        R_{\mathrm{ev}}={}&
        P_{\eta^\star,\mathrm{suf}}
          (\text{insufficient}\mid x)\\
        &+\tfrac12
        P_{\eta^\star,\mathrm{suf}}
          (\text{needs review}\mid x).
    \end{aligned}
    \label{eq:dr_evidence_risk}
\end{equation}
This quantity concerns whether the available evidence supports the
requested judgment; it is not a measure of physical disaster loss.

The highest-supported option outside the initial Top-2 is
\begin{equation}
    r=\arg\max_{k\in\mathcal K\setminus\{a,b\}}U_k.
    \label{eq:dr_rescue_candidate}
\end{equation}
A rescue candidate is activated only when $r$ has sufficient combined
support relative to $b$ and substantially stronger state agreement
than $a$:
\begin{equation}
    E_{\mathrm{rescue}}
    =\mathbb I\!\left[
        U_r\geq U_b+\delta_r
        \ \land\
        M_r\geq M_a+\delta_m
      \right].
    \label{eq:dr_rescue_signal}
\end{equation}
The review gate combines low Top-1 confidence, option--state conflict,
insufficient evidence, and candidate rescue:
\begin{equation}
    \begin{aligned}
        g=\mathbb I\!\bigl[{}&
        p_a<\tau_p\\
        &\lor(k_M\ne a
              \land M_{k_M}\geq M_a+\delta_v)\\
        &\lor R_{\mathrm{ev}}\geq\tau_{\mathrm{ev}}\\
        &\lor E_{\mathrm{rescue}}=1
        \bigr].
    \end{aligned}
    \label{eq:dr_routing_gate}
\end{equation}
All thresholds and margins are selected on validation data. When
$g=0$, DisasterRoute accepts $a$ directly; when $g=1$, it proceeds
to selective review.

\subsection{Constrained Selective Review}
\label{sec:method_selective_review}

Figure~\ref{fig:mach_overview}(c) shows the review branch.
Requesting review and expanding its candidate set are separate
decisions. For a routed example, the candidate set is
\begin{equation}
    \mathcal C(x)=
    \begin{cases}
        \{a,b,r\}, & E_{\mathrm{rescue}}=1,\\
        \{a,b\},   & E_{\mathrm{rescue}}=0.
    \end{cases}
    \label{eq:dr_candidate_set}
\end{equation}
Candidates retain their original A--E letters and option texts; they
are never renumbered. DisasterVL receives the original image,
question, and options in $\mathcal C(x)$, and independently re-encodes
the image. Predicted disaster-context fields, disputed option
attributes, and routing reasons are provided as cues to verify
against the image. When relevant, the reviewer examines the source
or trigger, motion or propagation, impact, and exposed objects,
rather than assuming that the state-head predictions are facts.

The reviewer produces a reasoning span in
\verb|<thinking>...</thinking>| and one original option letter in
\verb|<answering>...</answering>|. Let
$\widetilde y
=\Pi(D_\phi(I,q,\mathcal C(x),\widehat{\mathbf s}))$
be the answer returned by a strict parser $\Pi$. The final decision is
\begin{equation}
    \hat y=
    \begin{cases}
        \widetilde y,
        & g=1\ \text{and}\ \widetilde y\in\mathcal C(x),\\
        a, & \text{otherwise}.
    \end{cases}
    \label{eq:dr_final_decision}
\end{equation}
Thus, unreviewed examples return the System 1 prediction, while
invalid or out-of-set review outputs fall back to the same prediction.
The parser enforces output format and candidate membership; it does
not guarantee that a well-formed rationale is factually correct.

Let $T_{\mathrm{S1}}$, $T_{\mathrm{state}}$, $T_{\mathrm{gate}}$,
and $T_{\mathrm{review}}$ denote the respective computational costs.
Per-example inference time can be summarized as
\begin{equation}
    T(x)\approx
    T_{\mathrm{S1}}+T_{\mathrm{state}}
    +T_{\mathrm{gate}}+gT_{\mathrm{review}}.
    \label{eq:inference_cost}
\end{equation}
The state head reuses the System 1 representation, whereas
generative review is incurred only for routed examples. Reviewed
examples nevertheless require image re-encoding and autoregressive
decoding, and may therefore dominate mean and tail latency.

\begin{table*}[t]
    \centering

    \fontsize{9.0}{11.5}\selectfont
    \renewcommand{\arraystretch}{1.50}
    \setlength{\tabcolsep}{3.2pt}

    \begin{adjustbox}{max width=\textwidth}
    \begin{tabular}{@{}lcccccccccccc@{}}
        \toprule

        \textbf{Method}
        & \shortstack{\textbf{Correct}\\\textbf{/ Total}}
        & \shortstack{\textbf{Acc.}\\\textbf{(\%)} $\uparrow$}
        & \shortstack{\boldmath$\Delta$\textbf{Acc.}\\\textbf{(pp)}}
        & \textbf{Calls} $\downarrow$
        & \shortstack{\textbf{GCR}\\\textbf{(\%)} $\downarrow$}
        & \shortstack{\textbf{CC}\\\textbf{(\%)} $\uparrow$}
        & \shortstack{\textbf{Mean Lat.}\\\textbf{(ms)} $\downarrow$}
        & \shortstack{\textbf{P50}\\\textbf{(ms)} $\downarrow$}
        & \shortstack{\textbf{P95}\\\textbf{(ms)} $\downarrow$}
        & \shortstack{\textbf{Throughput}\\\textbf{(sample/s)} $\uparrow$}
        & \textbf{Speedup} $\uparrow$
        & \shortstack{\textbf{Peak Mem.}\\\textbf{(GiB)} $\downarrow$} \\

        \midrule
        \addlinespace[2.5pt]

        \textbf{DVL-AO}
        & 2167 / 2998
        & 72.28
        & \textemdash
        & 2998
        & 100.00
        & \textemdash
        & 278.69 $\pm$ 4.53
        & 184.35
        & 258.02
        & 3.59
        & 1.00$\times$
        & 4.40 \\

        \addlinespace[2.5pt]

        \textbf{DVL-AO-S}
        & 2137 / 2998
        & 71.28
        & $-1.00$
        & 2998
        & 100.00
        & \textemdash
        & 204.37 $\pm$ 2.84
        & 184.41
        & 259.26
        & 4.89
        & 1.36$\times$
        & 4.40 \\

        \addlinespace[4pt]

        \textbf{DVL-FR}
        & 1515 / 2998
        & 50.53
        & $-21.75$
        & 2998
        & 100.00
        & \textemdash
        & 3915.24 $\pm$ 21.07
        & 3954.11
        & 5352.40
        & 0.26
        & 0.07$\times$
        & 4.43 \\

        \addlinespace[2.5pt]

        \textbf{DVL-FR-S}
        & 1288 / 2998
        & 42.96
        & $-29.32$
        & 2998
        & 100.00
        & \textemdash
        & 3744.06 $\pm$ 12.75
        & 3900.39
        & 5291.31
        & 0.27
        & 0.07$\times$
        & 4.43 \\

        \addlinespace[4pt]

        \textbf{DVL-Logit}
        & 2166 / 2998
        & 72.25
        & $-0.03$
        & 0
        & 0.00
        & \textemdash
        & 149.47 $\pm$ 1.44
        & 144.20
        & 198.73
        & 6.69
        & 1.86$\times$
        & 4.38 \\

        \addlinespace[2.5pt]

        \textbf{S1-Direct}
        & 2332 / 2998
        & 77.79
        & $+5.50$
        & 0
        & 0.00
        & \textemdash
        & 150.05 $\pm$ 1.91
        & 144.45
        & 199.49
        & 6.66
        & 1.86$\times$
        & 4.37 \\

        \midrule
        \addlinespace[2.5pt]

        \textbf{DisasterRoute}
        & \textbf{2471 / 2998}
        & \textbf{82.42}
        & $\mathbf{+10.14}$
        & 710
        & 23.68
        & 93.52
        & 1215.01 $\pm$ 14.21
        & 141.21
        & 5634.79
        & 0.82
        & 0.23$\times$
        & 8.55 \\

        \addlinespace[2.5pt]
        \bottomrule
    \end{tabular}
    \end{adjustbox}

    \caption{Accuracy and inference efficiency on the DisasterBench test set. DVL-AO and DVL-AO-S use answer-only generation, while DVL-FR and DVL-FR-S generate reasoning and an answer. DVL-Logit reads DisasterVL's A--E option logits, whereas S1-Direct evaluates Disaster-S1 without routing or review; neither generates an answer sequence. DisasterRoute selectively invokes DisasterVL for System 2 review. Calls denotes the number of generative invocations, GCR denotes the fraction of questions requiring generation, and CC measures correct-answer coverage among reviewed candidate sets. $\Delta$Acc. and Speedup are relative to DVL-AO. Latency is measured with batch size one, and $\pm$ denotes the standard deviation of run means across three runs.}
    \label{tab:main_results}
    \vspace{-0.1cm}
\end{table*}

\begingroup
\vbadness=2000
\section{Experiments}

\subsection{Experimental Setup}
\label{sec:experimental_setup}

\paragraph{Dataset and Evaluation Protocol.}
We conduct experiments on DisasterBench, a benchmark for visual disaster question answering using low-altitude UAV imagery that covers hazard causation, propagation, impact, and response. The test split contains 2,998 image--question pairs, each with five A--E options. Disaster-S1 and the evidence-state head are fitted on the training split, while temperature scaling, routing parameters, and the standalone risk selector are fitted or selected using the validation split. All settings are fixed before testing, and test labels are used only for final evaluation. All evaluation protocols start from the same images, questions, and original options. Calibration and standalone risk-selection diagnostics are reported separately from the results of the complete DisasterRoute framework.

\paragraph{Implementation Details.}
Disaster-S1 is trained using language-side LoRA and five-way cross-entropy, with the vision encoder kept frozen, while the DisasterVL reviewer uses a fixed checkpoint. The main comparison uses NVIDIA L40S GPUs, BF16 precision, SDPA attention, and batch-size-one inference. Timing includes image decoding, preprocessing, model inference, routing, and answer parsing, but excludes model loading and warm-up. The main comparison is repeated three times, reporting mean latency and the standard deviation of run means. P50 and P95 are computed from pooled per-question latencies, while each general-purpose vision--language model baseline is evaluated in a single run. DisasterRoute's peak GPU memory measurement includes both resident models. Since model-sharding configurations may differ across runs, latency comparisons are not strictly paired.

\paragraph{Evaluation Metrics.}
We measure question-answering performance using accuracy (Acc.), with $\Delta$Acc. denoting the percentage-point difference from DVL-AO. Expected calibration error (ECE) and error-detection AUROC assess probability calibration and the ability of risk scores to identify initial prediction errors, respectively. For routing, routing rate (RR) is the fraction of questions selected, routed error rate (RER) is the initial error fraction within the selected subset, and direct-accept accuracy (DAA) measures accuracy on the unselected subset. For review, Calls and generative call rate (GCR) report the number of generative invocations and the fraction of questions requiring generation, respectively. Candidate coverage (CC) measures the fraction of reviewed questions whose candidate sets contain the correct answer, while Chg./Corr. records the numbers of changed predictions and corrected initial errors, respectively. Computational efficiency is assessed using mean latency, P50/P95 latency, throughput, speedup, and peak GPU memory. Throughput is the reciprocal of mean latency in seconds under serial inference, and speedup is the ratio of DVL-AO's mean latency to that of the evaluated method.

\subsection{Results and Analysis}
\label{sec:experimental_results}

\paragraph{Main Comparison.}
Table~\ref{tab:main_results} compares answer-only generation (DVL-AO and DVL-AO-S), full-reasoning generation (DVL-FR and DVL-FR-S), direct option readout (DVL-Logit and S1-Direct), and the complete DisasterRoute framework. The \texttt{-S} variants use shortened generation settings. DVL-Logit reads DisasterVL's A--E logits without decoding, whereas S1-Direct applies the same readout to Disaster-S1 without routing or review. The two direct protocols have similar mean latency (149.47 and 150.05~ms) but achieve 72.25\% and 77.79\% accuracy, respectively. This 5.54-percentage-point improvement supports the benefit of option-discrimination training for direct decisions. S1-Direct also exceeds DVL-AO's 72.28\% accuracy without autoregressive generation. The shortened generation variants reduce mean latency but also lower accuracy. Since prompts, generation limits, and output constraints differ across protocols, these comparisons do not isolate the effect of reasoning length.

DisasterRoute achieves 82.42\% accuracy, with 2,471 correct answers, compared with 2,332 for S1-Direct. This corresponds to a net gain of 139 correct answers, or 4.64 percentage points, and a 10.14-point improvement over DVL-AO. It invokes DisasterVL for 710 questions (23.68\% GCR), allowing the remaining 76.32\% to avoid generation; candidate coverage among reviewed questions is 93.52\%. Mean latency is 1215.01~ms, approximately 69\% lower than DVL-FR's 3915.24~ms, although P95 is slightly higher (5634.79 vs.\ 5352.40~ms). The complete framework remains slower on average than S1-Direct and increases peak GPU memory from 4.37 to 8.55~GiB. Consistent with Eq.~\eqref{eq:inference_cost}, routed questions still require image re-encoding and autoregressive review, while both models remain resident. Selective review therefore improves accuracy and reduces mean latency relative to uniform full reasoning, while retaining additional latency and memory costs relative to direct inference.

\begin{table}[!t]
    \centering
    \normalfont
    \fontsize{9.38}{12.2}\selectfont
    \renewcommand{\arraystretch}{1.10}
    \setlength{\tabcolsep}{1pt}

    \begin{tabular*}{\columnwidth}{
        @{\extracolsep{\fill}}lccccc@{}
    }
        \toprule

        & \multicolumn{1}{c}{\textbf{Calibration}}
        & \multicolumn{4}{c}{\textbf{Risk Selection}} \\[1.5pt]

        \cmidrule(lr){2-2}
        \cmidrule(lr){3-6}

        \textbf{Config.}
        & \shortstack{\textbf{ECE}\\$\downarrow$}
        & \shortstack{\textbf{AUROC}\\$\uparrow$}
        & \shortstack{\textbf{RR}\\\textbf{(\%)} $\downarrow$}
        & \shortstack{\textbf{RER}\\\textbf{(\%)} $\uparrow$}
        & \shortstack{\textbf{DAA}\\\textbf{(\%)} $\uparrow$}
        \\[1pt]

        \midrule
        \addlinespace[0.5pt]

        \textbf{Raw+LC}
        & 0.16
        & 0.8362
        & 13.61
        & 71.08
        & 85.48 \\

       \addlinespace[0.5pt]

        \textbf{TS+LC}
        & \textbf{0.03}
        & 0.8348
        & 14.74
        & 72.40
        & \textbf{86.46} \\

       \addlinespace[0.5pt]

        \textbf{TS+RS}
        & \textbf{0.03}
        & \textbf{0.8453}
        & \textbf{10.51}
        & \textbf{76.19}
        & 84.12 \\

       \addlinespace[0.5pt]
        \bottomrule
    \end{tabular*}

    \caption{Confidence calibration and standalone risk selection for Disaster-S1. Thresholds are selected on validation data. For TS+RS, AUROC evaluates the error score before disaster-consequence weighting, while RR, RER, and DAA describe the resulting selection. These diagnostics are separate from the complete DisasterRoute gate.}
    \label{tab:mach_calibration}
    \vspace{-0.1cm}
\end{table}

\paragraph{Disaster-S1 Calibration and Risk Selection.}
Table~\ref{tab:mach_calibration} evaluates confidence calibration and standalone risk selection for Disaster-S1. Raw+LC selects questions with low uncalibrated Top-1 confidence, while TS+LC applies temperature scaling before selection. TS+RS instead uses a fitted risk selector with disaster-consequence weighting. For TS+RS, weighting affects selection, whereas AUROC evaluates the underlying error score before weighting. These configurations use validation-selected operating points and are evaluated separately from the complete DisasterRoute gate in Table~\ref{tab:mach_routing}. The standalone LC thresholds are also selected independently of the LC threshold used by that gate.

Temperature scaling reduces ECE from 0.16 to 0.03, while AUROC changes from 0.8362 to 0.8348, indicating that improved calibration does not necessarily improve error ranking. TS+LC achieves the highest DAA (86.46\%) at a routing rate of 14.74\%, with 72.40\% RER. TS+RS achieves the highest AUROC (0.8453) and RER (76.19\%) while selecting 10.51\% of questions, with 84.12\% DAA. It therefore selects a smaller subset with greater initial error concentration, whereas TS+LC selects more initially incorrect questions in absolute terms. These results characterize different operating points rather than establish superiority at an equal review budget.

\begin{table}[!t]
    \centering
    \normalfont
    \fontsize{9.38}{12.2}\selectfont
    \renewcommand{\arraystretch}{1.10}
    \setlength{\tabcolsep}{1pt}

    \begin{tabular*}{\columnwidth}{
        @{\extracolsep{\fill}}lccc@{}
    }
        \toprule

        & \multicolumn{1}{c}{\textbf{Triggered Subset}}
        & \multicolumn{2}{c}{\textbf{Routing Quality}} \\[1.5pt]

        \cmidrule(lr){2-2}
        \cmidrule(lr){3-4}

        \rule{0pt}{2.8ex}
        \textbf{Trigger}
        & \textbf{Samples}
        & \textbf{RER (\%)} $\uparrow$
        & \textbf{CC (\%)} $\uparrow$ \\[1pt]

        \midrule
       \addlinespace[0.5pt]

        \textbf{LC}
        & 380
        & 75.00
        & 92.63 \\

       \addlinespace[0.5pt]

        \textbf{OSC}
        & 334
        & \textbf{100.00}
        & \textbf{97.31} \\

       \addlinespace[0.5pt]

        \textbf{IE}
        & 122
        & 91.80
        & 90.16 \\

       \addlinespace[0.5pt]

        \textbf{CR}
        & 294
        & \textbf{100.00}
        & 89.12 \\

       \addlinespace[0.5pt]
        \bottomrule
    \end{tabular*}

    \caption{Evidence-guided routing triggers in DisasterRoute. LC, OSC, IE, and CR denote low confidence, option--state conflict, insufficient evidence, and candidate rescue, respectively. RER measures the initial error rate within each triggered subset, while CC measures correct-answer coverage in the corresponding review candidate sets. Triggered subsets may overlap.}
    \label{tab:mach_routing}
    \vspace{-0.1cm}
\end{table}

\paragraph{Evidence-Guided Routing.}
Table~\ref{tab:mach_routing} reports the four routing triggers in DisasterRoute: low confidence (LC), option--state conflict (OSC), insufficient evidence (IE), and candidate rescue (CR). CR corresponds to $E_{\mathrm{rescue}}=1$, which triggers review and adds a candidate outside the initial Top-2 when the combined-support and state-agreement conditions are satisfied. LC flags 380 questions with 75.00\% RER, while IE flags 122 with 91.80\% RER. OSC and CR both attain 100.00\% RER, but their candidate coverage differs (97.31\% vs.\ 89.12\%), showing that concentrating initial errors does not guarantee retaining the correct answer for review. The 294 CR cases constitute the Top-2+R branch in Table~\ref{tab:mach_verification}. Since multiple conditions can trigger for the same question, the subsets overlap, and their union comprises the 710 questions sent to DisasterVL for review.

\begin{table}[!t]
    \centering
    \normalfont
    \fontsize{9.38}{12.2}\selectfont
    \renewcommand{\arraystretch}{1.10}
    \setlength{\tabcolsep}{2pt}

    \begin{tabular*}{\columnwidth}{
        @{\extracolsep{\fill}}lcccc@{}
    }
        \toprule

        & &
        \multicolumn{1}{c}{
            \shortstack{\textbf{Candidate}\\\textbf{Set}}
        }
        & \multicolumn{2}{c}{\textbf{Review}} \\[1.5pt]

        \cmidrule(lr){3-3}
        \cmidrule(lr){4-5}

        \textbf{Branch}
        & \textbf{Samples}
        & \shortstack{\textbf{CC (\%)}\\$\uparrow$}
        & \shortstack{\textbf{Chg./}\\\textbf{Corr.}}
        & \shortstack{\textbf{Final Acc.}\\\textbf{(\%)}}
        \\[1pt]

        \midrule
       \addlinespace[0.5pt]

        \textbf{Direct}
        & 2288
        & \textemdash
        & \textemdash
        & 97.77 \\

       \addlinespace[0.5pt]

        \textbf{Top-2}
        & 416
        & 96.63
        & 129 / 124
        & 51.68 \\

       \addlinespace[0.5pt]

        \textbf{Top-2+R}
        & 294
        & 89.12
        & 28 / 19
        & 6.46 \\

        \midrule
        \addlinespace[2.0pt]

        \textbf{All reviewed}
        & 710
        & 93.52
        & 157 / 143
        & 32.96 \\

       \addlinespace[0.5pt]
        \bottomrule
    \end{tabular*}

    \caption{Direct acceptance and constrained candidate review in
    DisasterRoute. Top-2+R adds one rescue candidate outside the initial
    Top-2 when $E_{\mathrm{rescue}}=1$. CC denotes candidate coverage;
    Chg./Corr. counts changed predictions and corrected initial errors
    relative to Disaster-S1. Final accuracy is computed within each subset.}
    \label{tab:mach_verification}
    \vspace{-0.1cm}
\end{table}

\paragraph{Constrained Candidate Review.}
Table~\ref{tab:mach_verification} breaks down DisasterRoute into direct acceptance and two candidate-review branches: Top-2 and Top-2+R. The 2,288 directly accepted questions achieve 97.77\% accuracy. For the 710 questions routed to DisasterVL, candidate coverage reaches 93.52\%, while final accuracy is 32.96\%. Review changes 157 initial predictions and corrects 143 initial errors, yielding a net gain of 139 correct answers over S1-Direct and an overall accuracy of 82.42\%. The reviewed questions form a selectively routed subset in which the gate concentrates initial prediction errors.

The Top-2 branch reviews 416 questions using the two highest-scoring Disaster-S1 options, achieving 96.63\% CC and 51.68\% final accuracy, with 129 changed predictions and 124 corrected errors. When $E_{\mathrm{rescue}}=1$, the Top-2+R branch adds one rescue candidate outside the initial Top-2. This branch contains 294 questions and achieves 89.12\% CC and 6.46\% final accuracy, with 28 changed predictions and 19 corrected errors. Since the two branches contain different selectively routed questions, their accuracy difference does not measure the causal effect of adding a rescue candidate. The gap between candidate coverage and final accuracy further shows that retaining the correct answer in the candidate set is insufficient to ensure successful review.

\begin{table}[t]
    \centering
    \normalfont
    \fontsize{9.38}{12.2}\selectfont
    \renewcommand{\arraystretch}{1.10}
    \setlength{\tabcolsep}{2pt}

    \begin{tabular*}{\columnwidth}{
        @{\extracolsep{\fill}}lcccc@{}
    }
        \toprule

        \textbf{Method}
        & \shortstack{\textbf{Acc.}\\\textbf{(\%)} $\uparrow$}
        & \shortstack{\textbf{GCR}\\\textbf{(\%)} $\downarrow$}
        & \shortstack{\textbf{Mean Lat.}\\\textbf{(s)} $\downarrow$}
        & \shortstack{\textbf{P95}\\\textbf{(s)} $\downarrow$} \\

        \midrule
       \addlinespace[0.5pt]

        InternVL3.5-1B
        & 32.92
        & 100.00
        & 4.23
        & 6.41 \\

       \addlinespace[0.5pt]

        InternVL3.5-2B
        & 55.34
        & 100.00
        & 6.29
        & 17.21 \\

       \addlinespace[0.5pt]

        Qwen2.5-VL-3B
        & 46.00
        & 100.00
        & 5.27
        & 9.65 \\

        \midrule
       \addlinespace[0.5pt]

        DVL-FR
        & 50.53
        & 100.00
        & 3.92
        & 5.35 \\

       \addlinespace[0.5pt]

        S1-Direct
        & 77.79
        & \textbf{0.00}
        & \textbf{0.15}
        & \textbf{0.20} \\

       \addlinespace[0.5pt]

        \textbf{DisasterRoute}
        & \textbf{82.42}
        & 23.68
        & 1.22
        & 5.63 \\

       \addlinespace[0.5pt]
        \bottomrule
    \end{tabular*}

    \caption{Comparison with general-purpose VLMs on DisasterBench.
    The three general-purpose baselines and DVL-FR generate reasoning
    and an answer for every question. GCR denotes the percentage of
    questions requiring autoregressive generation.
    Each general-purpose baseline is evaluated in one run;
    DVL-FR, S1-Direct, and DisasterRoute results are taken from the
    three-run main comparison.}
    \label{tab:base_vlm_comparison}
    \vspace{-0.1cm}
\end{table}

\paragraph{Comparison with General-Purpose VLMs.}
Table~\ref{tab:base_vlm_comparison} compares three general-purpose VLMs evaluated with reasoning and answer generation against DVL-FR, S1-Direct, and DisasterRoute. Among the three general-purpose baselines, InternVL3.5-2B achieves the highest accuracy of 55.34\%, with a mean latency of 6.29~s. S1-Direct achieves 77.79\% accuracy at 0.15~s without autoregressive generation, while DisasterRoute reaches 82.42\% at 1.22~s, invoking DisasterVL for review on 23.68\% of questions. DVL-FR, which generates reasoning and an answer for every question, achieves 50.53\% accuracy at 3.92~s. DisasterRoute has a lower mean latency than DVL-FR, although its P95 latency is slightly higher (5.63 vs.\ 5.35~s), indicating that selective review still incurs substantial tail latency. The three general-purpose baselines are each evaluated in one run, while the remaining results come from the three-run main comparison. These latency comparisons are descriptive and do not constitute strictly paired measurements.

\begin{figure}[t]
    \centering
    \includegraphics[width=\columnwidth]{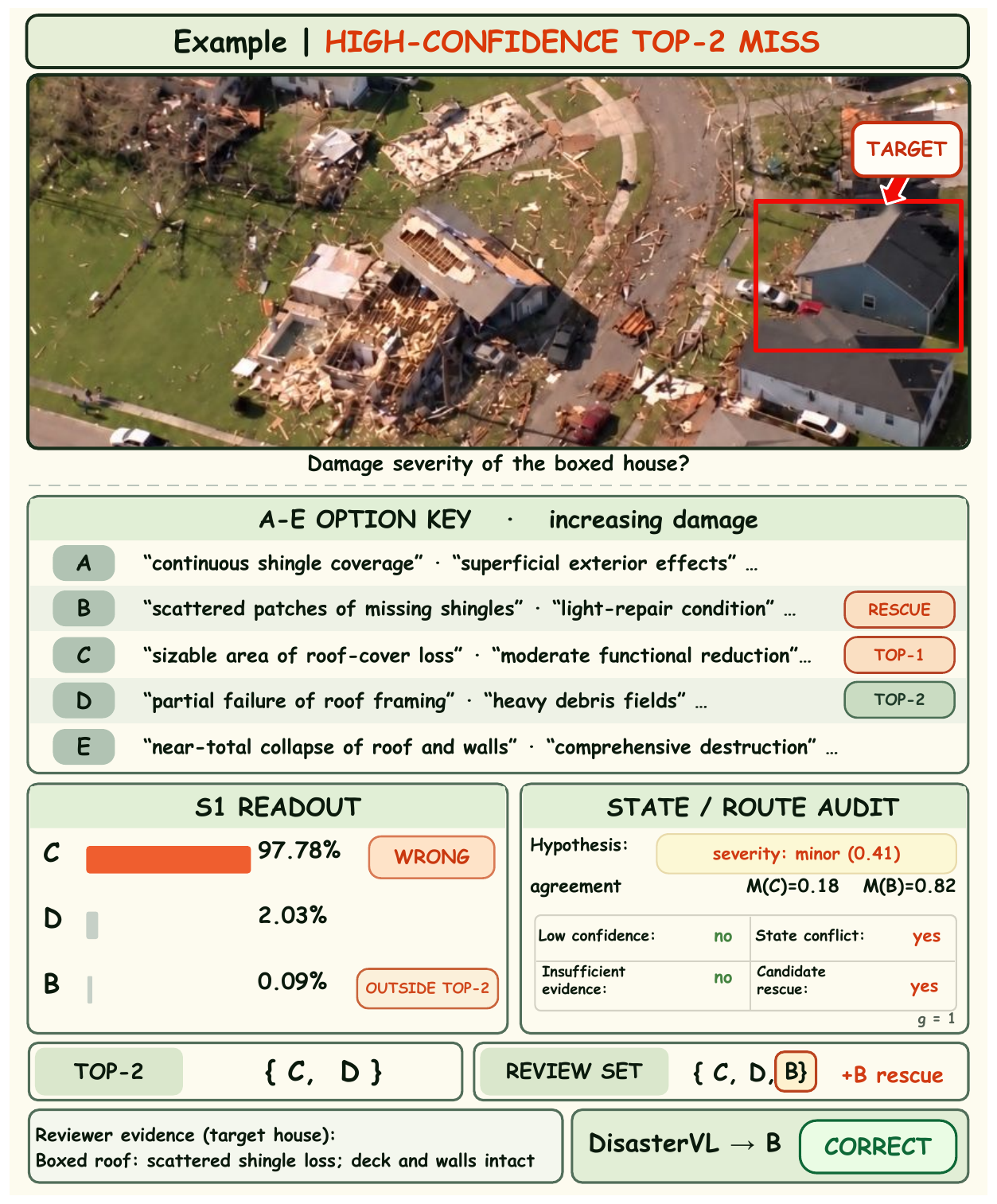}
    \caption{Schematic illustration of candidate rescue. The top panel shows the target house and summaries of options A--E; the lower panels sequentially present Disaster-S1's option probabilities, state and routing checks, the candidate set expansion from $\{C,D\}$ to $\{C,D,B\}$, and DisasterVL's review of the target house.}
    \label{fig:example}
    \vspace{-0.2cm}
\end{figure}

\paragraph{Illustrative Candidate Rescue.}
Figure~\ref{fig:example} illustrates a high-confidence initial error and omission of the correct answer from the Top-2 candidates through a damage-severity assessment of the boxed house. Disaster-S1 assigns 97.78\% probability to C, while the evidence-state head predicts minor damage with a highest class probability of 0.41; the option--state agreement score for C is only 0.18. Thus, the option--state conflict triggers review even though the low-confidence condition is inactive. The initial Top-2 set $\{C,D\}$ excludes the correct option B, which is subsequently added through candidate rescue. DisasterVL then re-examines the boxed house using the candidate set $\{C,D,B\}$: scattered shingle loss, together with an intact roof deck and walls, supports the final selection of B. This example illustrates the distinct roles of conflict-triggered review, candidate rescue in expanding the available choices, and visual review in making the final judgment.

\endgroup

\section{Conclusion}

We presented DisasterRoute, an evidence-guided dual-system framework for UAV-based visual disaster question answering. Its fast decision model, Disaster-S1, is trained for option discrimination and produces five-option probabilities in a single multimodal forward pass without autoregressive generation. DisasterRoute combines prediction confidence, question-specific evidence sufficiency, and option--state agreement to select questions for generative review by DisasterVL, with conditional candidate rescue expanding the review set beyond the initial Top-2. On DisasterBench, option-discrimination training improves direct decision accuracy, and selective review further raises accuracy from 77.79\% to 82.42\% while invoking generation for 23.68\% of questions. The framework reduces mean latency by approximately 69\% relative to applying DisasterVL's full reasoning to every question, although it incurs additional latency and memory costs over direct inference. High candidate coverage does not ensure successful review, and tail latency remains a limitation. These findings support selective coordination of direct decisions and generative review for disaster question answering. Future work will improve evidence estimation and candidate comparison, reduce review costs, and evaluate generalization across disaster settings.

\begingroup
\hbadness=2500
\bibliography{custom}

@inproceedings{Kuckreja2024GeoChat,
  author    = {Kartik Kuckreja and Muhammad Sohail Danish and Muzammal Naseer and Abhijit Das and Salman Khan and Fahad Shahbaz Khan},
  title     = {{GeoChat}: Grounded Large Vision-Language Model for Remote Sensing},
  booktitle = {Proceedings of the IEEE/CVF Conference on Computer Vision and Pattern Recognition},
  pages     = {27831--27840},
  year      = {2024},
  doi       = {10.1109/CVPR52733.2024.02629},
  url       = {https://openaccess.thecvf.com/content/CVPR2024/html/Kuckreja_GeoChat_Grounded_Large_Vision-Language_Model_for_Remote_Sensing_CVPR_2024_paper.html}
}

@inproceedings{Muhtar2025LHRSBot,
  author    = {Dilxat Muhtar and Zhenshi Li and Feng Gu and Xueliang Zhang and Pengfeng Xiao},
  title     = {{LHRS-Bot}: Empowering Remote Sensing with {VGI}-Enhanced Large Multimodal Language Model},
  booktitle = {Computer Vision -- ECCV 2024},
  series    = {Lecture Notes in Computer Science},
  volume    = {15132},
  pages     = {440--457},
  publisher = {Springer},
  year      = {2025},
  doi       = {10.1007/978-3-031-72904-1_26},
  url       = {https://link.springer.com/chapter/10.1007/978-3-031-72904-1_26}
}

@inproceedings{Soni2025EarthDial,
  author    = {Sagar Soni and Akshay Dudhane and Hiyam Debary and Mustansar Fiaz and Muhammad Akhtar Munir and Muhammad Sohail Danish and Paolo Fraccaro and Campbell D. Watson and Levente J. Klein and Fahad Shahbaz Khan and Salman Khan},
  title     = {{EarthDial}: Turning Multi-sensory Earth Observations to Interactive Dialogues},
  booktitle = {Proceedings of the IEEE/CVF Conference on Computer Vision and Pattern Recognition},
  pages     = {14303--14313},
  year      = {2025},
  url       = {https://openaccess.thecvf.com/content/CVPR2025/html/Soni_EarthDial_Turning_Multi-sensory_Earth_Observations_to_Interactive_Dialogues_CVPR_2025_paper.html}
}

@inproceedings{Revankar2025MONITRS,
  author    = {Shreelekha Revankar and Utkarsh Mall and Cheng Perng Phoo and Kavita Bala and Bharath Hariharan},
  title     = {{MONITRS}: Multimodal Observations of Natural Incidents Through Remote Sensing},
  booktitle = {Advances in Neural Information Processing Systems},
  volume    = {38},
  year      = {2025},
  doi       = {10.52202/085713-4355},
  url       = {https://papers.nips.cc/paper_files/paper/2025/hash/bd96a50dfd2314e48787581840a07a1a-Abstract-Datasets_and_Benchmarks_Track.html}
}

@inproceedings{Chen2025RSCC,
  author    = {Zhenyuan Chen and Chenxi Wang and Ningyu Zhang and Feng Zhang},
  title     = {{RSCC}: A Large-Scale Remote Sensing Change Caption Dataset for Disaster Events},
  booktitle = {Advances in Neural Information Processing Systems},
  volume    = {38},
  year      = {2025},
  doi       = {10.52202/085713-2290},
  url       = {https://papers.nips.cc/paper_files/paper/2025/hash/62867024377cac4233195949b9db0ebd-Abstract-Datasets_and_Benchmarks_Track.html}
}

@inproceedings{Wang2025DisasterM3,
  author    = {Junjue Wang and Weihao Xuan and Heli Qi and Zhihao Liu and Kunyi Liu and Yuhan Wu and Hongruixuan Chen and Jian Song and Junshi Xia and Zhuo Zheng and Naoto Yokoya},
  title     = {{DisasterM3}: A Remote Sensing Vision-Language Dataset for Disaster Damage Assessment and Response},
  booktitle = {Advances in Neural Information Processing Systems},
  volume    = {38},
  year      = {2025},
  doi       = {10.52202/085713-5398},
  url       = {https://papers.nips.cc/paper_files/paper/2025/hash/ec80d18205e39d42a27192d5f3ddd688-Abstract-Datasets_and_Benchmarks_Track.html}
}

@misc{Zhang2026DisasterBench,
  author        = {Tan Zhang and Quanyou Li and Lu Zhang and Jun Liu and Xiaofeng Zhu and Ping Hu},
  title         = {{DisasterBench}: A Multimodal Benchmark for {UAV}-Based Disaster Response in Complex Environments},
  year          = {2026},
  eprint        = {2606.06217},
  archivePrefix = {arXiv},
  primaryClass  = {cs.CV},
  doi           = {10.48550/arXiv.2606.06217},
  url           = {https://arxiv.org/abs/2606.06217}
}

@misc{Almeida2026Jev,
  author       = {Diogo Almeida},
  title        = {Introducing System One Models \& {Jev}},
  howpublished = {TypeSafe AI Blog},
  year         = {2026},
  month        = sep,
  url          = {https://typesafe.ai/blog/introducing-system-one-models-and-jev}
}

@misc{Deusser2026JevBenchmark,
  author        = {Tobias Deu{\ss}er and Lorenz Sparrenberg and Rafet Sifa},
  title         = {Evaluating and Benchmarking the System One Model {Jev}},
  year          = {2026},
  eprint        = {2609.37647},
  archivePrefix = {arXiv},
  primaryClass  = {cs.CL},
  doi           = {10.48550/arXiv.2609.37647},
  url           = {https://arxiv.org/abs/2609.37647}
}

@misc{Yu2026VisualJev,
  author        = {Guanxu Yu and Yuhang Yao},
  title         = {Visual {Jev}: Accurate and Efficient Decisions from Shared Visual Context},
  year          = {2026},
  eprint        = {2609.25845},
  archivePrefix = {arXiv},
  primaryClass  = {cs.CV},
  doi           = {10.48550/arXiv.2609.25845},
  url           = {https://arxiv.org/abs/2609.25845}
}

@misc{Li2026LLMasJev,
  author        = {Yinheng Li and Justin Wagle},
  title         = {{LLM-as-Jev}: {LLMs} Are Already {Jev}-Style Decision Models -- When and How to Fine-Tune Them},
  year          = {2026},
  eprint        = {2610.02076},
  archivePrefix = {arXiv},
  primaryClass  = {cs.CL},
  doi           = {10.48550/arXiv.2610.02076},
  url           = {https://arxiv.org/abs/2610.02076}
}

@inproceedings{Whitehead2022ReliableVQA,
  author    = {Spencer Whitehead and Suzanne Petryk and Vedaad Shakib and Joseph E. Gonzalez and Trevor Darrell and Anna Rohrbach and Marcus Rohrbach},
  title     = {Reliable Visual Question Answering: Abstain Rather Than Answer Incorrectly},
  booktitle = {Computer Vision -- ECCV 2022},
  series    = {Lecture Notes in Computer Science},
  volume    = {13696},
  pages     = {148--166},
  publisher = {Springer},
  year      = {2022},
  doi       = {10.1007/978-3-031-20059-5_9},
  url       = {https://link.springer.com/chapter/10.1007/978-3-031-20059-5_9}
}

@inproceedings{Dancette2023SelectiveVQA,
  author    = {Corentin Dancette and Spencer Whitehead and Rishabh Maheshwary and Ramakrishna Vedantam and Stefan Scherer and Xinlei Chen and Matthieu Cord and Marcus Rohrbach},
  title     = {Improving Selective Visual Question Answering by Learning From Your Peers},
  booktitle = {Proceedings of the IEEE/CVF Conference on Computer Vision and Pattern Recognition},
  pages     = {24049--24059},
  year      = {2023},
  url       = {https://openaccess.thecvf.com/content/CVPR2023/html/Dancette_Improving_Selective_Visual_Question_Answering_by_Learning_From_Your_Peers_CVPR_2023_paper.html}
}

@article{Dille2026HazardContinuum,
  author  = {Dille, Antoine and Dewitte, Olivier and Broeckx, Jente
             and Verbist, Koen and Sindiso Dube, Andile
             and Poesen, Jean and Vanmaercke, Matthias},
  title   = {Capturing the Complete Landslide--Debris-Rich Flood
             Continuum for Accurate Inventory, Susceptibility and
             Exposure Mapping -- Lessons from Cyclone Idai},
  journal = {Natural Hazards and Earth System Sciences},
  volume  = {26},
  pages   = {2561--2577},
  year    = {2026},
  doi     = {10.5194/nhess-26-2561-2026},
  url     = {https://nhess.copernicus.org/articles/26/2561/2026/}
}

@article{Kyrkou2020EmergencyNet,
  author  = {Christos Kyrkou and Theocharis Theocharides},
  title   = {{EmergencyNet}: Efficient Aerial Image Classification for Drone-Based Emergency Monitoring Using Atrous Convolutional Feature Fusion},
  journal = {IEEE Journal of Selected Topics in Applied Earth Observations and Remote Sensing},
  volume  = {13},
  pages   = {1687--1699},
  year    = {2020},
  doi     = {10.1109/JSTARS.2020.2969809},
  url     = {https://doi.org/10.1109/JSTARS.2020.2969809}
}

@inproceedings{Leviathan2023SpeculativeDecoding,
  author    = {Yaniv Leviathan and Matan Kalman and Yossi Matias},
  title     = {Fast Inference from Transformers via Speculative Decoding},
  booktitle = {Proceedings of the 40th International Conference on Machine Learning},
  series    = {Proceedings of Machine Learning Research},
  volume    = {202},
  pages     = {19274--19286},
  publisher = {PMLR},
  year      = {2023},
  url       = {https://proceedings.mlr.press/v202/leviathan23a.html}
}

@article{Evans2013DualProcess,
  author  = {Jonathan St. B. T. Evans and Keith E. Stanovich},
  title   = {Dual-Process Theories of Higher Cognition:
             Advancing the Debate},
  journal = {Perspectives on Psychological Science},
  volume  = {8},
  number  = {3},
  pages   = {223--241},
  year    = {2013},
  doi     = {10.1177/1745691612460685},
  url     = {https://doi.org/10.1177/1745691612460685}
}

@inproceedings{Saha2025System1x,
  author    = {Swarnadeep Saha and Archiki Prasad and Justin Chen
               and Peter Hase and Elias Stengel-Eskin
               and Mohit Bansal},
  title     = {{System 1.x}: Learning to Balance Fast and Slow
               Planning with Language Models},
  booktitle = {International Conference on Learning Representations},
  year      = {2025},
  url       = {https://proceedings.iclr.cc/paper_files/paper/2025/hash/605e59e78284907aa4fce5280838def3-Abstract-Conference.html}
}

@inproceedings{Sun2025FastSlow,
  author    = {Guangyan Sun and Mingyu Jin and Zhenting Wang
               and Chenglong Wang and Siqi Ma and Qifan Wang
               and Tong Geng and Yingnian Wu and Yongfeng Zhang
               and Dongfang Liu},
  title     = {Visual Agents as Fast and Slow Thinkers},
  booktitle = {International Conference on Learning Representations},
  year      = {2025},
  url       = {https://proceedings.iclr.cc/paper_files/paper/2025/hash/14fb70b97c70d5cc5445cd2d5bf818db-Abstract-Conference.html}
}

@misc{Zhou2026PixelJev,
  author        = {Xunlan Zhou and Xianliang Yang and Li Zhao},
  title         = {From Text Decisions to Pixels: An Study of {Jev}-Style Visual Choice Model},
  year          = {2026},
  eprint        = {2609.29283},
  archivePrefix = {arXiv},
  primaryClass  = {cs.AI},
  doi           = {10.48550/arXiv.2609.29283},
  url           = {https://arxiv.org/abs/2609.29283}
}

@misc{Duan2026JEVal,
  author        = {Feiyu Duan and Jiayu Lin and Jia Wang and Jun Xiang
                   and Jialiang Wu and Xinnong Zhang and Hanqi Yan
                   and Siyuan Wang and Zhongyu Wei},
  title         = {General Decision Models: Benchmarking and Insights Beyond {Jev}},
  year          = {2026},
  eprint        = {2610.03935},
  archivePrefix = {arXiv},
  primaryClass  = {cs.CL},
  doi           = {10.48550/arXiv.2610.03935},
  url           = {https://arxiv.org/abs/2610.03935}
}

@misc{Li2026JevJudge,
  author        = {Yubo Li and Yidi Miao and Ramayya Krishnan and Rema Padman},
  title         = {{JEV}-as-a-Judge: Accept When Confident, Escalate When Unsure},
  year          = {2026},
  eprint        = {2609.26550},
  archivePrefix = {arXiv},
  primaryClass  = {cs.AI},
  doi           = {10.48550/arXiv.2609.26550},
  url           = {https://arxiv.org/abs/2609.26550}
}

@misc{Wu2026ReflexJev,
  author        = {Tiantong Wu and Wei Yang Bryan Lim},
  title         = {{REFLEX} with {Jev} for Efficient Selective Control in {LLM} Agents},
  year          = {2026},
  eprint        = {2609.26532},
  archivePrefix = {arXiv},
  primaryClass  = {cs.AI},
  doi           = {10.48550/arXiv.2609.26532},
  url           = {https://arxiv.org/abs/2609.26532}
}
\endgroup

\end{document}